\documentclass[letterpaper]{article} 
\usepackage{aaai25}  
\usepackage{times}  
\usepackage{helvet}  
\usepackage{courier}  
\usepackage[hyphens]{url}  
\usepackage{graphicx} 
\usepackage{subcaption}
\usepackage{pdfpages}
\usepackage{booktabs}
\usepackage{tablefootnote}
\usepackage{natbib}  
\usepackage{caption} 
\usepackage{algorithm}
\usepackage{algorithmic}
\usepackage{xcolor}

\usepackage{newfloat}
\usepackage{listings}
\DeclareCaptionStyle{ruled}{labelfont=normalfont,labelsep=colon,strut=off} 
\floatstyle{ruled}
\newfloat{listing}{tb}{lst}{}
\floatname{listing}{Listing}
\usepackage{enumerate}

\title{BiMi Sheets: Infosheets for bias mitigation methods}
\author{
    MaryBeth Defrance,
    Guillaume Bied,
    Maarten Buyl,
    Jefrey Lijffijt,
    Tijl De Bie
}
\affiliations{
    Ghent University\\

    Technologiepark-Zwijnaarde 116, \\
    9052 Gent, Belgium\\
    \{firstname\}.\{lastname\}@ugent.be
}

\usepackage{bibentry}

\definecolor{MyForestGreen}{rgb}{0.110, 0.451, 0.110}

\definecolor{MyGoldenrod}{rgb}{0.855, 0.647, 0.125}

\definecolor{MyTangerine}{rgb}{0.850, 0.475, 0.000}

\definecolor{MyBurgundy}{rgb}{0.800, 0.0, 0.125}

\begin{document}

\maketitle

\begin{abstract}
Over the past 15 years, hundreds of bias mitigation methods have been proposed in the pursuit of fairness in machine learning (ML). However, algorithmic biases are domain-, task-, and model-specific, leading to a `portability trap': bias mitigation solutions in one context may not be appropriate in another. Thus, a myriad of design choices have to be made when creating a bias mitigation method, such as the formalization of fairness it pursues, and where and how it intervenes in the ML pipeline. This creates challenges in benchmarking and comparing the relative merits of different bias mitigation methods, and limits their uptake by practitioners.

We propose BiMi Sheets as a portable, uniform guide to document the design choices of any bias mitigation method. This enables researchers and practitioners to quickly learn its main characteristics and to compare with their desiderata. Furthermore, the sheets' structure allow for the creation of a structured database of bias mitigation methods. In order to foster the sheets' adoption, we provide a platform for finding and creating BiMi Sheets at bimisheet.com. 

\end{abstract}

\section{Introduction}

The risk of algorithms reproducing or amplifying social biases has drawn considerable attention to fairness in AI. While technical solutions may not always be adequate \cite{wachter2021fairness}, hundreds of algorithmic bias mitigation methods have nevertheless been proposed in the pursuit of fair machine learning.
The diversity of these methods stems from a large number of necessary design choices, and from the complex nature of fairness itself. For instance, methods may differ in (i) how fairness is formally defined, (ii) the dataset types they are compatible with, (iii) the machine learning tasks they target, (iv) the stage of the machine learning pipeline where they intervene, (v) the machine learning models they are compatible with, and more. Despite being all ostensibly designed with the same goal--mitigating algorithmic bias--they thus generalize poorly across socio-technical contexts. This difficulty was dubbed the `portability trap' by \citet{Selbst2019}.

Yet, this diversity of bias mitigation methods raises challenges--both for their adoption in practice and for the progress of academic research. On the one hand, practitioners report challenges when searching for bias mitigation methods appropriate for their use case. Based on a user study, \citet{Deng2022} find that the adoption of fairness tools would improve if \textit{more
guidance and support in contextualizing toolkit functionalities or outputs, beyond the level of documentation provided by standard software packages} was provided. On the other hand, good benchmarking practices have been key to progress in various sub-fields of machine learning. Yet, in fair-ML, novel methods are often benchmarked against a small, established set of bias mitigation methods (most of which proposed between 2016 and 2019), ignoring more recent advances in the literature, and even though some of the baseline methods may not be viable alternatives in practice
\cite{Delaney_Fu_Wachter_Mittelstadt_Russell_2024, Han_Chi_Chen_Wang_Zhao_Zou_Hu_2024, cruz2024unprocessing}. 

In the end, such challenges may result in an increasing misalignment between the technical tools for mitigating algorithmic bias and the socio-technical concerns that motivated the tackling of this bias in the first place \cite{weerts2024can}. 
We contend that steps must be taken to elaborate on the technical details and assumptions for bias mitigation methods in a consistent manner, such that the technical scope of each method can be clarified. 
In turn, a consistent common language for bias mitigation tools may then facilitate the socio-technical discussion on the appropriateness of a given method for a particular use case. 

\begin{figure*}[]
  \centering
  \fbox{
  \includegraphics[width=0.8\linewidth, trim={3.5cm 6cm 3.5cm 3.5cm}]{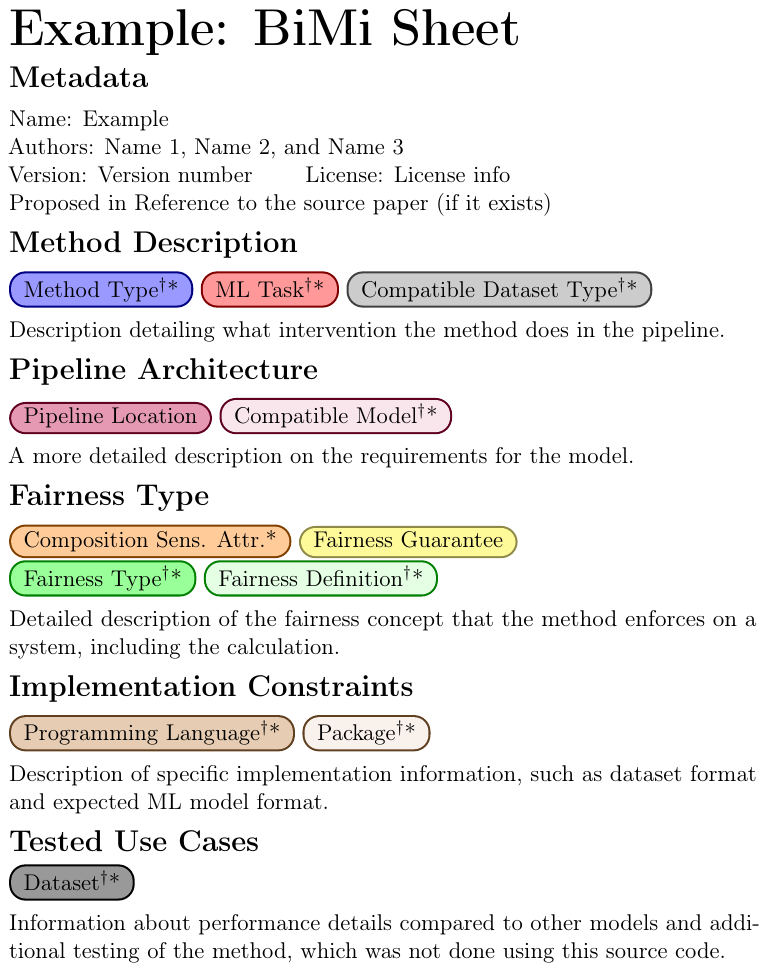}} 
  
  \caption{Bare-bones example of a BiMi Sheet. The symbol \textsuperscript{\textdagger} denotes that the label in the tag is free to choose. The symbol * denotes that multiple tags for this property can be provided in a BiMiSheet.}
  \label{fig:bimi_example}
  \end{figure*}




\paragraph{Contributions} We therefore introduce \emph{BiMi Sheets}: information sheets for bias mitigation methods with two goals. 
\begin{enumerate}
    \item To serve as \emph{documentation}, providing detailed, consistent information to practitioners on the bias mitigation method's design choices and constraints
    \item To establish \emph{structure} for the myriad of existing bias mitigation methods such that research and benchmarks can find common ground and continue to progress.
\end{enumerate}
\noindent The sheets' design is informed by a variety of taxonomies and terminologies from surveys on AI fairness, as well as key needs from practioners and researchers identified in previous research. A schematic example is shown in Fig.~\ref{fig:bimi_example}. The BiMi Sheets consist of six sections that respectively detail the bias mitigation method, its location in an ML pipeline, its formalization of fairness, other implementation constraints, reports on tested use cases, and any additional metadata. To kickstart the adoption of BiMi Sheets, we already provide 24 BiMi Sheets for popular, recent bias mitigation methods on a searchable web application at \url{bimisheet.com}.



\paragraph{Outline}
The rest of this paper is organized as follows. After reviewing related work, we highlight the existence of documentation debt for bias mitigation methods, motivating the proposal of BiMi Sheets.
We proceed to describe the proposed structure of a BiMi Sheet, motivating the adopted design choices based on the needs of users and common practices within AI fairness research. 
We end with a discussion on the ongoing challenges for the documentation and structuring of bias mitigation methods in the future.

\section{Related work}
\label{sec:related_work}

Our work is informed by past efforts on improving the \emph{documentation} and \emph{structuring} of bias mitigation methods.


\subsection{Documentation for fairness in AI}
\label{sec:documentation}
Documentation of bias mitigation methods remains understudied. However, some documentation initiatives exist for communicating fairness characteristics of resources, other than bias mitigation methods. We limit our discussion to works focussing specifically on the fairness aspects and therefore not discuss initiatives like System Cards \cite{Gursoy_Kakadiaris_2022} or Audit Cards \cite{Staufer_Yang_Reuel_Casper_2025}.  

\paragraph{Datasheets for datasets \cite{Gebru_Morgenstern2021}} Datasheets for datasets call for the documentation of the creation, composition, intended uses, maintenance, and other properties of datasets. 
The goal of datasheets is two-fold.
First, to ensure the proper use of datasets by making the critical knowledge of domain experts available through the datasheet, which quickly communicates the capabilities, strengths, and limitations of the dataset to the machine learning expert. 
Second, to urge the creators to critically reflect on the limitations and (un)intended impact of their dataset by pushing them to answer a diverse set of questions.

\paragraph{Model cards for models \cite{Mitchell_2019}} Model cards are short documents accompanying trained machine learning models that provide benchmarked evaluation in a variety of conditions. Model cards are widely in use, as they are integrated for models provided on HuggingFace \cite{HuggingFace2025}. The main elements discuss its intended use, the sensitive attributes the model can take into account, its performance metrics, training data, evaluation data, and limitations. These let users determine the trained model's applicability for their use case and enable a performance comparison with other trained models. 


\paragraph{FactSheets by IBM \cite{Arnold2019}} Factsheets focus on documented AI services. This notion differs slightly from the model cards. An AI service could consist out of a single AI model, but it can also be constructed of many elements. FactSheets also aim to encompass more than just the fairness aspect of the service, including also explainability and security. The concept `AI service' is very broad, so the suggested structure of the FactSheets is inevitably rather large in order to encompass all conceivable characteristics that such a system might exhibit.

\paragraph{Aequitas \cite{Saleiro_Kuester_Hinkson_London_Stevens_Anisfeld_Rodolfa_Ghani_2019}} The Aequitas Toolkit aims at auditing machine learning models on the possible biases they may exhibit and provide guidance for determining the desired properties of a fair machine learning model. This initiative focusses on creating documentation for a specific application rather than a method. 

\paragraph{BiMi Sheets in comparison} Datasheets, Model cards, and FactSheets document the biases present in the dataset, model, and AI service respectively. These biases in part arise from real-world biases. BiMi Sheets are distinct in this regard, as they do not document biases, but the design choices of bias mitigation methods. Indeed, a bias mitigation method's strengths and weaknesses stem from the design choices made when creating the algorithm. These design choices translate into enforcing a specific fairness notion \cite{Green_2019}. 

\subsection{Structuring for fairness in AI}
\label{sec:surveys}

BiMi Sheets complement surveys of bias mitigation methods in machine learning, and build on the taxonomies they propose to categorize these methods. While \citet{Mehrabi_Morstatter_Saxena_Lerman_Galstyan_2021, Caton_Haas_2024, Pagano_2023} review fair machine learning with an emphasis on binary classification, other works focus on specific domains such as machine learning on graphs \cite{Laclau_Largeron_Choudhary_2024}, large language models \cite{Gallegos_Rossi_Barrow_Tanjim_Kim_Dernoncourt_Yu_Zhang_Ahmed_2024, Chu_Wang_Zhang_2024},
information access and recommendation \cite{Ekstrand_2022, Li_al_2023}, or gender bias in natural language processing and computer vision \cite{Bartl_Mandal_Leavy_Little_2024}. Beyond their apparent diversity, these surveys leverage common elements to describe bias mitigation methods--a method's associated fairness types and definitions, compatible dataset types, associated machine learning tasks, pipeline, and model compatibility. BiMi Sheets build on these common elements and on domain-specific taxonomies to propose a flexible documentation structure for bias mitigation methods across domains.

\section{Documentation debt in bias mitigation}
\label{sec:challenges}
\begin{table*}[t]
\centering
\begin{tabular}{p{5.5cm}| c c c c c c c c}
 & \rotatebox{75}{Method Approach} & \rotatebox{75}{Compatible Models} & \rotatebox{75}{Pipeline Location} & \rotatebox{75}{Compatible Datasets} & \rotatebox{75}{\parbox{3.0cm}{Composition \\ Sensitive Attributes \footnotemark}} &  \rotatebox{75}{Fairness Guarantee} & \rotatebox{75}{Fairness Notion}  &  \rotatebox{75}{\parbox{2.9cm}{Implementation \\ Constraints}}    \\ \hline
 Adversarial Debiasing - AIF360\footnotemark & \color{MyForestGreen} A & \color{MyForestGreen}A\color{MyTangerine}* &  \color{MyForestGreen}A & \color{MyForestGreen}A\color{MyTangerine}* & \color{MyForestGreen}A & \color{MyForestGreen}A\color{MyTangerine}* & \color{MyForestGreen}A & \color{MyGoldenrod}N \\
 Calibrated Equalized Odds - AIF360 & \color{MyBurgundy}? & \color{MyForestGreen}A\color{MyTangerine}* & \color{MyForestGreen}A & - & \color{MyBurgundy}? & \color{MyBurgundy}? & \color{MyForestGreen}A & \color{MyForestGreen}A \\
 Deterministic Reranking - AIF360 & \color{MyGoldenrod}N & \color{MyGoldenrod}N\color{MyTangerine}* & \color{MyForestGreen}A & - & \color{MyBurgundy}? & \color{MyGoldenrod}N\color{MyTangerine}* & \color{MyGoldenrod}N\color{MyTangerine}* & \color{MyForestGreen}A\\
 Disparate Impact Remover - AIF360 & \color{MyBurgundy}? & - & \color{MyForestGreen}A & \color{MyForestGreen}A\color{MyTangerine}* & \color{MyBurgundy}? & \color{MyForestGreen}A\color{MyTangerine}* &  \color{MyBurgundy}? & \color{MyForestGreen}A\\
 Exponent. Gradient Reduct. - AIF360 & \color{MyForestGreen}A & - & \color{MyForestGreen}A & - & \color{MyBurgundy}? & \color{MyForestGreen}A & \color{MyForestGreen}A & \color{MyForestGreen}A \\
 GerryFair Classifier - AIF360 & \color{MyBurgundy}? & \color{MyForestGreen}A & \color{MyForestGreen}A & - & \color{MyBurgundy}? & \color{MyGoldenrod}N\color{MyTangerine}* & \color{MyForestGreen}A & \color{MyForestGreen}A, \color{MyGoldenrod}N\\
 Grid Search Reduction - AIF360 & \color{MyForestGreen}A & - & \color{MyForestGreen}A & - & \color{MyBurgundy}? & \color{MyBurgundy}? & \color{MyForestGreen}A & \color{MyForestGreen}A \\
 LFR - AIF360 & \color{MyForestGreen}A & - & \color{MyForestGreen}A & \color{MyBurgundy}? & \color{MyBurgundy}? & \color{MyForestGreen}A\color{MyTangerine}* & \color{MyForestGreen}A & \color{MyForestGreen}A \\
 MetaFair Classifier - AIF360 & \color{MyBurgundy}? & \color{MyBurgundy}? & \color{MyForestGreen}A & \color{MyBurgundy}? & \color{MyForestGreen}A & \color{MyForestGreen}A\color{MyTangerine}* & \color{MyForestGreen}A, \color{MyGoldenrod}N & \color{MyForestGreen}A  \\
 Optimized Preprocessing - AIF360 & \color{MyForestGreen}A & - & \color{MyForestGreen}A & \color{MyBurgundy}? & \color{MyBurgundy}? & \color{MyBurgundy}? & \color{MyForestGreen}A & \color{MyForestGreen}A\\

 Prejudice Remover - AIF360 & \color{MyForestGreen}A & \color{MyForestGreen}A\color{MyTangerine}* & \color{MyForestGreen}A & - & \color{MyBurgundy}? & \color{MyForestGreen}A\color{MyTangerine}* & \color{MyBurgundy}? & \color{MyForestGreen}A  \\
 
 Reject Option Classification - AIF360 & \color{MyForestGreen}A & \color{MyForestGreen}A\color{MyTangerine}* & \color{MyForestGreen}A & - & \color{MyBurgundy}? & \color{MyBurgundy}? & \color{MyForestGreen}A & \color{MyForestGreen}A \\
 Reweighing - AIF360 & \color{MyBurgundy}? & \color{MyForestGreen}A\color{MyTangerine}* & \color{MyForestGreen}A & - & \color{MyBurgundy}? & \color{MyBurgundy}? & \color{MyBurgundy}? & \color{MyForestGreen}A   \\
 Adversarial Mitigation - Fairlearn & \color{MyForestGreen}U & \color{MyForestGreen}U & \color{MyForestGreen}U\color{MyTangerine}* & \color{MyForestGreen}U\color{MyTangerine}* & \color{MyForestGreen}U & \color{MyForestGreen}A\color{MyTangerine}* & \color{MyForestGreen}U & \color{MyForestGreen}U \\
 Correlation Remover - Fairlearn & \color{MyForestGreen}U & - & \color{MyForestGreen}U & \color{MyForestGreen}U\color{MyTangerine}* & \color{MyForestGreen}A & \color{MyForestGreen}U, \color{MyForestGreen}A & \color{MyForestGreen}U & \color{MyForestGreen}U\color{MyTangerine}* \\
 Reductions - Fairlearn & \color{MyForestGreen}U & \color{MyForestGreen}U & \color{MyForestGreen}U\color{MyTangerine}* & - & \color{MyBurgundy}? & \color{MyForestGreen}A & \color{MyForestGreen}U & \color{MyForestGreen}A \\
 Threshold Optimizer - Fairlearn & \color{MyForestGreen}U & \color{MyForestGreen}U\color{MyTangerine}* & \color{MyForestGreen}U & - & \color{MyBurgundy}? & \color{MyForestGreen}U & \color{MyForestGreen}U & \color{MyForestGreen}U\\
 Error-Parity & \color{MyForestGreen}U & \color{MyForestGreen}U & \color{MyForestGreen}U & - & \color{MyGoldenrod}N\color{MyTangerine}* & \color{MyForestGreen}U\color{MyTangerine}* & \color{MyForestGreen}U & \color{MyForestGreen}U \\
 Fairret & \color{MyForestGreen}U\color{MyTangerine}* & \color{MyForestGreen}U\color{MyTangerine}* & \color{MyForestGreen}U\color{MyTangerine}* & \color{MyBurgundy}? & \color{MyGoldenrod}N\color{MyTangerine}* & \color{MyGoldenrod}N\color{MyTangerine}* & \color{MyForestGreen}A\color{MyTangerine}* & \color{MyForestGreen}U \\
 OxonFair & \color{MyForestGreen}U & \color{MyForestGreen}U\color{MyTangerine}* & \color{MyForestGreen}U & - & \color{MyForestGreen}U & \color{MyForestGreen}U & \color{MyForestGreen}U & \color{MyForestGreen}U \\
\end{tabular}
\caption{Occurrence for where different characteristics of a bias mitigation methods can be found in online documentation.  \textcolor{MyForestGreen}{A}: API reference, \textcolor{MyForestGreen}{U}: User documentation, \textcolor{MyGoldenrod}{N}: Notebooks, \textcolor{MyBurgundy}{?}: Not found, -: Not relevant for the method. \textcolor{MyTangerine}*: The information can be deduced by someone with an advanced knowledge in AI fairness. This table only states whether or not this information is available, not how straightforward this information is to find.}
\label{table:occurence_characteristics_online_docs}
\end{table*}


Our work is a response to the \emph{documentation debt} in the current state of bias mitigation methods that limits their comparability in research and their usability in practice. In what follows, we revisit the motivation for documentation for researchers and practioners, before mapping the current state of documentation across recent and popular toolkits.

\paragraph{Documentation for researchers} Researchers that propose bias mitigation methods need a way to compare their method against existing methods such that its strengths and weaknesses can be analyzed on a quantitative and qualitative basis. Yet, unlike other fields in ML where progress can steadily be mentioned on benchmarks and leaderboards, no general leaderboard can exist for bias mitigation methods \cite{Wang_Hertzmann_Russakovsky_2024} because formalizations of algorithmic bias are so specific to the socio-technical context. Hence, the design choices of a method, and the resulting constraints for the socio-technical context, must be alignable across methods for their comparison to be meaningful \cite{Defrance_Buyl_Bie_2024}.

\paragraph{Documentation for practitioners} In contrast, practitioners want methods that are easily integrated into their existing ML pipelines without imposing invasive requirements \cite{Lee2021}. Yet, practioners have complained that documentation of bias mitigation methods is often incomplete or overly technical \cite{Deng2022}, while they struggle to find methods that are compatible with their ML task and the datasets they are working with \cite{Richardson2021}. 

\footnotetext[1]{We assume that each bias mitigation method can use binary sensitive attributes. We note a question mark if it is not clearly stated which composition it can handle and only examples using binary sensitive attributes are provided.}

\footnotetext[2]{In AIF360 the API reference and User documentation are identical, we therefore only report the API reference.}

\paragraph{The current state of documentation} Having established the importance of documentation for both researchers and practitioners, let us now consider if and how the online documentation of bias mitigation methods sufficiently address their needs. Focusing on recent and well-known fairness toolkits, we consulted the documentation of AIF360 \cite{aif360_2018}, FairLearn \cite{weerts2023fairlearn}, error-parity \cite{cruz2024unprocessing}, fairret \cite{Buyl_Defrance_Bie_2024}, and  OxonFair \cite{Delaney_Fu_Wachter_Mittelstadt_Russell_2024}. For each, we checked the documentation along several axes, including
\begin{enumerate}[(i)]
    \item a description of the method's overall approach
    \item the underlying ML models it is compatible with (e.g. logistic regression, neural nets, etc.)
    \item the stage of the ML pipeline where it intervenes
    \item the types of compatible datasets
    \item how it delineates protected groups
    \item whether it offers guarantees on fairness
    \item how it defines fairness mathematically
    \item any implementation constraints (e.g. the need for a \textit{scikit-learn} pipeline)
\end{enumerate}

Table~\ref{table:occurence_characteristics_online_docs} documents if, and where, selected characteristics of bias mitigation methods can be found in their online documentation. We find no uniformity to exist with respect to where or even whether these elements are communicated in online documentation across, or even within, libraries. Additionally, many characteristics are left implicit to some extent, requiring deduction based on advanced knowledge of bias mitigation methods. Altogether, the importance of documenting various aspects of bias mitigation methods for different types of users, and the existence of a \textit{documentation debt}, motivate the introduction of BiMi Sheets as a portable and uniform guide to document bias mitigation methods.

\section{Structure of the sheets}\label{sec:structure}
BiMi Sheets document the specific implementation of a bias mitigation method. Any algorithmic intervention aimed at reducing bias, either somewhere in the model pipeline or in the outcome, is considered a bias mitigation method. Accordingly, BiMi Sheets need to account for a wide range of methods. 

BiMi Sheets are structured in six sections, each of which discusses a specific property of the bias mitigation method. All sections, except the one for metadata, contain both labels and free text. The labels provide a structuring of the bias mitigation methods, making their comparison easier. The free text serves to communicate the intricate details which define the specific method. 

For the sake of illustration, a bare-bones version of a BiMi Sheet can be found in Figure~\ref{fig:bimi_example}. The names for all labels can be found in this bare-bones BiMi Sheet. Remark the shading of these labels: lighter-shaded labels are conceptually children components of the normally shaded version. 

In the following subsections we discuss each section of the BiMi Sheet. We first provide a narrative description which situates the content of the section with the literature. Each subsection is concluded with a list discussing the label information, such as possible values. 


\subsection{Metadata}
The Metadata section communicates basic non-fairness related information for the bias mitigation method.

\paragraph{Label Information}
\begin{itemize}
 \item \textbf{Name and Authors.} The combination of the bias mitigation method and authors' names identify a specific method. The authors are the creators of the implementation, not the concept. The authors can refer to a set of individuals or to the fairness toolkit directly.
 
 \item \textbf{Version.} The version conveys the method's software version number. A BiMi Sheet corresponds to a specific version as other versions might have a different set of capabilities, or approach fairness slightly differently. 
 
 \item \textbf{License.} License information is important for practitioners: An unknown software license complicates adoption.
 
 \item \textbf{Proposed in.} If the method was first proposed in a research paper, then refer to the paper here.
\end{itemize}

\subsection{Method description}


Several taxonomies have been proposed in order to provide an initial structuring of bias mitigation methods \cite{Bartl_Mandal_Leavy_Little_2024, Caton_Haas_2024, Chu_Wang_Zhang_2024}. This structuring is based on the type of intervention a method undertakes on the pipeline. The type of intervention is a design choice that corresponds to a fairness notion. We rename the labels from these taxonomies to \textbf{Method Type}. A well-known example in binary classification of such a method type is \textit{Adversarial Debiasing}. With this method type some element in the pipeline is transformed in order to prevent an adversarial model from guessing a sample's sensitive attributes.

The need for fairness interventions stretches a broad range of applications. As a consequence, bias mitigation methods cannot function for all applications. We characterize an application as the combination of two labels: \textbf{ML Task} and \textbf{Compatible~Dataset~Type}. The ML Task stands for the objective the ML model aims to achieve. Practitioners have critiqued the large focus on classification in proposed bias mitigation methods \cite{Richardson2021}. The ML Task and Dataset Type labels will facilitate finding methods for other purposes and perhaps entice researchers to create bias mitigation methods for a broader range of problems. 

Both labels are needed to illustrate the application in order to account for pre- and post-processing methods. The nature of pre- and post-processing methods often causes one of the aforementioned labels to become trivial. Pre-processing methods mitigate biases in the dataset themselves, meaning that these methods generally can be used for any ML Task compatible with a specific dataset type. For example, pre-processing methods for graphs can be seen as \textit{task independent}, however its possible application is communicated through its compatible dataset type, \textit{graphs} \cite{Laclau_Largeron_Choudhary_2024}. On the other hand, post-processing methods affect the model outputs and become independent of the dataset type. A post-processing method for binary classification can be used for both tabular, image, and text datasets \cite{Delaney_Fu_Wachter_Mittelstadt_Russell_2024}.

A bias mitigation method might focus on a specific sub-problem within an ML task. For instance, binary classification can either be evaluated with respect to hard labels or soft scoring. Bias mitigation methods designed for hard labels suffer in performance when evaluated for soft scoring compared to methods designed for that purpose \cite{Defrance_Buyl_Bie_2024}. The reported depth within a ML task depends on the bias mitigation method's properties and focus.

\paragraph{Label Information}

\begin{table*}
\begin{tabular}{p{3.4cm} l p{10.7cm}}
\toprule
Source & Pipeline location & Method Types \\
\midrule
\citet{Caton_Haas_2024} - \qquad Binary Classification &  Pre-processing & Blinding, Causal Methods, Sampling and Subgroup Analysis, Transformation,  Relabelling and Perturbation, Reweighing, Adversarial Learning \\
& In-processing & Transformation, Reweighing, Regularization and Constraint Optimisation, Adversarial Learning, Bandits \\
& Post-processing & Calibration, Thresholding \\ \\

\citet{Bartl_Mandal_Leavy_Little_2024} -  & Pre-processing & Sampling  \\
Computer Vision & In-processing & Adversarial
Debiasing \\
& Intra-processing & Adversarial Debiasing, Learning Representations, Model fine-tuning\\
& Post-processing & Model fine-tuning \\ \\

\citet{Chu_Wang_Zhang_2024} -  & Pre-processing & Data Augmentation, Prompt Tuning\\
Large Language Models & In-processing & Loss Function Modification, Auxiliary Module\\
& Intra-processing & Model Editing, Decoding Method Modification\\
& Post-processing & Chain of Thought, Rewriting \\

\citet{Gallegos_Rossi_Barrow_Tanjim_Kim_Dernoncourt_Yu_Zhang_Ahmed_2024} - LLMs & Pre-processing & Data Balancing, Selective Replacement, Interpolation, Dataset Filtering, Instance Reweighting, Equalized Teacher Model Probabilities, Exemplary examples, Word Lists, Modified Prompting Language, Control Tokens, Continuous Prompt Tuning, Projection-based Mitigation \\
& In-processing & Architecture Modification, Equalizing Objectives, Fair Embeddings, Attention, Predicted token distribution, Dropout, Contrastive Learning, Adversarial Learning, Reinforcement Learning, Selective Parameter Updating, Filtering Model Parameters \\
& Intra-processing & Constrained Next-token Search, Modified Token Distribution, Weight Redistribution, Modular Debiasing Networks \\
& Post-processing & Keyword Replacement, Machine Translation, Other Neural Rewriters \\
\bottomrule
\end{tabular}
\caption{Overview of the proposed Method Types from the surveys of \citet{Caton_Haas_2024, Bartl_Mandal_Leavy_Little_2024, Chu_Wang_Zhang_2024}.}
\label{tab:method_types}
\end{table*}

\begin{table*}
\begin{tabular}{l l p{9.6cm}}
\toprule
Source & Context & ML Tasks \\
\midrule
\cite{Bartl_Mandal_Leavy_Little_2024} & NLP & Occupation Classification, Sentiment Analysis, Machine Translation
 \\

\cite{Mehrabi_Morstatter_Saxena_Lerman_Galstyan_2021} & General & Classification, Regression, Community detection, Clustering, Machine translation, Semantic role labeling, Named Entity Recognition \\

\cite{Laclau_Largeron_Choudhary_2024} & Graphs & Node classification, Edge prediction, Community detection, Graph property prediction \\

\cite{Gallegos_Rossi_Barrow_Tanjim_Kim_Dernoncourt_Yu_Zhang_Ahmed_2024} & LLMs & Classification, Question-answering, Logical reasoning, Fact retrieval, Information extraction \\
\bottomrule

\end{tabular}
\caption{Overview of discussed ML tasks in the surveys of \citet{Bartl_Mandal_Leavy_Little_2024, Mehrabi_Morstatter_Saxena_Lerman_Galstyan_2021, Laclau_Largeron_Choudhary_2024, Gallegos_Rossi_Barrow_Tanjim_Kim_Dernoncourt_Yu_Zhang_Ahmed_2024}.}
\label{tab:MLTasks}

\end{table*}

\begin{itemize}
	\item \textbf{Method Type.} Method type provides information on the type of intervention the method does in the pipeline. These method types and their characteristics can be found in surveys discussing AI fairness. An overview of method types can be found in Table~\ref{tab:method_types}. 
	
	\item \textbf{ML Task.} ML Task communicates the specific machine learning tasks for which the method was designed. It is possible to have multiple ML task labels for one method. Some examples of ML tasks are given in Table~\ref{tab:MLTasks}.
	
	\item \textbf{Compatible Dataset Type.} The compatible dataset labels communicate the dataset types for which the method is known to work. Examples of dataset types include tabular datasets, image datasets, text datasets, and recommendation datasets. 

	\item \textbf{Method Description.} This section provides a detailed explanation of the intervention that the bias mitigation method does in the pipeline. This explanation should be unique compared to other BiMi Sheets, unless a specific method has several implementations. 
\end{itemize}

\subsection{Pipeline architecture}

One of the most used properties to differentiate bias mitigation methods is their \textbf{Pipeline Location}, i.e. in which part of the machine learning pipeline the intervention occurs. This influences the capabilities of a method \cite{Defrance_Buyl_Bie_2024}. 
Additionally, pipeline location is an important compatibility constraint in practical settings \cite{Richardson2021}, since it is not always possible to intervene at any location. Historically, the prevailing division was pre-, in-, and post-processing \cite{Bartl_Mandal_Leavy_Little_2024, Caton_Haas_2024, Ekstrand_2022, Laclau_Largeron_Choudhary_2024,Mehrabi_Morstatter_Saxena_Lerman_Galstyan_2021}. Novel research on large language models introduced a fourth pipeline 
location, namely intra-processing \cite{Chu_Wang_Zhang_2024, Gallegos_Rossi_Barrow_Tanjim_Kim_Dernoncourt_Yu_Zhang_Ahmed_2024}.  

Research on the needs of practitioners with regards to fairness toolkits has shown that an important feature of a toolkit is how easily it integrates in existing workflows \cite{Deng2022, Lee2021}. Knowing the \textbf{Compatible Models} is an important constraint determining whether or not the method would fit in the existing workflow.
Pre-processing methods are often method independent as their intervention occurs before data is processed by the model. Even so, Reweighing---a pre-processing method from AIF360 \cite{aif360_2018}---requires a model that can incorporate sample weights. Post-processing methods process the output of a model. Therefore, the compatibility of a model is not dependent on the technique, but on the output format. This results in most post-processing classification methods requiring probabilistic classifiers. 

More information on the constraints imposed by the method on the pipeline is described in the free text of this section. This additional information could for example list the models with which a model independent method has been tested. Another possible use is for mentioning particular assumptions, such as for Deterministic Reranking from AIF360 \cite{aif360_2018}, which assumes that the provided rankings are ordered by descending score. 

\paragraph{Label Information}

\begin{table*}
\begin{tabular}{l p{14.6cm}}
\toprule
Location & Description \\
\midrule
Pre-processing & Pre-processing techniques try to transform the data so the underlying discrimination is removed \cite{dAlessandro_2017}. This data includes both training data and prompts \cite{Chu_Wang_Zhang_2024}.\\
In-processing & In-processing techniques try to modify and change state-of-the-art learning
algorithms to remove discrimination during the model training process \cite{dAlessandro_2017}. This includes
making modifications to the optimization process by adjusting the loss function and incorporating auxiliary modules \cite{Chu_Wang_Zhang_2024}.\\
Intra-processing & The Intra-processing focuses on mitigating bias in pretrained or fine-tuned models during the inference stage without requiring additional training \cite{Chu_Wang_Zhang_2024}.\\
Post-processing & Post-processing approaches modify the results generated by
the model to mitigate biases \cite{Chu_Wang_Zhang_2024}.\\
\bottomrule

\end{tabular}
\caption{Descriptions of pipeline locations by \citet{Chu_Wang_Zhang_2024, dAlessandro_2017}.}
\label{tab:pipeline_loc}

\end{table*} 

\begin{itemize}
	\item \textbf{Pipeline Location.} The location in the pipeline signifies where the method intervenes. This property affects the capabilities of the bias mitigation method. An overview of the possible pipeline locations and their description can be found in Table~\ref{tab:pipeline_loc}.
	\item \textbf{Compatible Model.} States the models which are technically compatible and have been shown to have acceptable performance after applying the bias mitigation method. 
	\item \textbf{Model description.} In the model description, more detailed information is provided on the theoretically compatible models. For example, the pre-processing method CorrelationRemover from Fairlearn \cite{weerts2023fairlearn} is model independent, however it has been shown that it is most appropriate to be combined with linear models. This is an example of additional information that should be available when assessing the compatibility with a use case. 
\end{itemize}

\subsection{Fairness type}

Bias occurs when people with differing sensitive attributes are treated differently. Most older bias mitigation methods can only account for binary attributes, with gender as prototypical example. However, this is often insufficient to mitigate biases in practice. Today, it is understood that there is a need to account for more complex sensitive attributes, including \textbf{Compositions of Sensitive Attributes}. For example, the works of \citet{Buolamwini_Gebru_2018} call for intersectional fairness, meaning that each combination of sensitive attributes denotes one sensitive group. This can be translated to one categorical attribute which contains this concatenation of sensitive attributes. 

A problem arises when many sensitive attributes are present or when certain combinations are rare. Due to a lack of statistical power for small groups, it might be impossible to calculate a fairness measure for each group. An alternative solution is to account for parallel attributes, meaning that per set of sensitive attributes fairness is ensured \cite{Defrance_Buyl_Bie_2024}. This is a weaker fairness notion compared to intersectional fairness, but still accounts for multiple axes of sensitive attributes. 

Fairness in AI is often described as satisfying a specific fairness definition. While such a \textbf{Fairness Guarantee} of satisfying a definition may seem desirable, providing a guarantee is often accompanied by direct implications on other performance metrics.
\citet{Sylvester_Raff_2018} have found that practitioners would sometimes prefer applying a method that improves fairness, but does not fully guarantee fairness in order to achieve their design goals.

Several interpretations of fairness exist in literature. Bias mitigation methods target one or more of these interpretations. We call these interpretations \textbf{Fairness Types} to emphasize the link with the specific \textbf{Fairness Definitions} that fall under these higher-order fairness types. Binary classification has a well-known split based on fairness type, namely between \textit{Group Fairness} and \textit{Individual Fairness}. Other fairness concepts exist such as egalitarianism, Rawlsian justice, or Nozick's entitlement theory, however no practical bias mitigation methods have been proposed for these concepts. Still, the flexibility of the labels would allow for these concepts to be stated as a fairness type.

The previously mentioned labels cannot convey the full fairness notion that a bias mitigation method enforces: a \textbf{Fairness Description} text area is therefore provided to allow for further details. This description aims to help against the `portability trap' of \citet{Selbst2019}. The `portability trap' discusses how enforcing fairness is highly contextual and can only be reused after an informed decision. The fairness description provides this detailed description on which compatibility with a new case can be gauged. 

\paragraph{Label Information}

\begin{itemize}
	\item \textbf{Composition Sensitive Attributes.} This label signifies which combinations of sensitive attributes can be handled by the method. An overview of common compositions can be found in Table~\ref{tab:comp_sens_atrr}. Binary attributes are a subset of categorical attributes and categorical attributes are a subset of parallel attributes. 
	\item \textbf{Fairness Guarantee.} The fairness guarantee states to what degree a fairness method can guarantee fairness. The three possible types of fairness guarantee for a method can be found in Table~\ref{tab:fairness_guarantee}. A method that provides fairness guarantees fails if the chosen fairness constraint cannot be achieved.
	\item \textbf{Fairness Type.} The fairness type denotes the fairness interpretation that a method can satisfy. Depending on the context different fairness types exist. An example of fairness types can be found in Table~\ref{tab:fairness_types}. 
	\item \textbf{Fairness Definition.} Fairness definitions are the mathematical formulas a method can enforce on a pipeline. A fairness definition is associated with a fairness type. A vast assortment of fairness definitions have been proposed in the literature. We refer to the surveys of \citet{Caton_Haas_2024, Laclau_Largeron_Choudhary_2024}, and \citet{Gallegos_Rossi_Barrow_Tanjim_Kim_Dernoncourt_Yu_Zhang_Ahmed_2024} for examples of fairness definitions.
	\item \textbf{Fairness Description} The fairness description provides a more detailed explanation how fairness is envisioned and enforced in the method. This includes the mathematical approach for calculating the fairness difference. It also provides the necessary information surrounding the parameters for tuning fairness. If there are any constraints or relaxations with regards to the sensitive attributes, this must be noted here.
\end{itemize}

\begin{table*}
\centering
\begin{tabular}{l p{13.5cm}}
\toprule
Composition & Description \\
\midrule
Binary Attribute & The method can only take into account one binary attribute whether an individual belongs to a specific group. \\
Categorical Attributes & The method can encode categorical attributes. This means that it there are a certain number of groups for which an individual can only belong into one of these groups. \\
Parallel Attributes & The method can handle a set of categorical attributes, where it aims to enforce fairness along the axis of every categorical attribute.\\
Numerical Attribute & The method can handle continuous valued attributes.\\
\bottomrule
\end{tabular}
\caption{Overview of Compositions of Sensitive Attributes.}
\label{tab:comp_sens_atrr}
\end{table*}

\begin{table*}
\begin{tabular}{l p{13cm}}
\toprule 
Fairness Guarantee & Description \\
\midrule
Fairness Guaranteed & The method can guarantee the satisfaction of a fairness constraint. This fairness constraint is often configurable through the method's hyperparameters. \\
Tunable Fairness Strength & The method provides a hyperparameter where the strength of the fairness intervention relative to other performance such as performance can be set. \\
No Fairness Guarantee & The method's fairness intervention strength cannot be tuned and no formal fairness guarantee is provided by the method. \\
\bottomrule
\end{tabular}
\caption{Overview of the possible Fairness Guarantee a method could provide.}
\label{tab:fairness_guarantee}
\end{table*}

\begin{table*}
\begin{tabular}{l p {13.4cm}}
\toprule
Context & Existing fairness types \\
\midrule
Binary classification & Group Fairness, Individual Fairness, Subgroup Fairness \cite{Mehrabi_Morstatter_Saxena_Lerman_Galstyan_2021} \\ 
& Group Fairness, Individual Fairness, Counterfactual Fairness \cite{Caton_Haas_2024} \\
Large Language Models & Group Fairness, Individual Fairness \cite{Chu_Wang_Zhang_2024} \\
& Embedding-based Fairness, Probability-based Fairness, and Generated Text-based Fairness \cite{Gallegos_Rossi_Barrow_Tanjim_Kim_Dernoncourt_Yu_Zhang_Ahmed_2024} \\
Recommender Systems & Consumer Fairness, Provider Fairness \cite{Ekstrand_2022} \\
Graphs & Structural Metrics, Representation Fairness, Fair Prediction \cite{Laclau_Largeron_Choudhary_2024} \\
\bottomrule
\end{tabular}
\caption{Examples of fairness types.}
\label{tab:fairness_types}
\end{table*}

\subsection{Implementation}
BiMi Sheets focus on a specific implementation of a bias mitigation method, an important feature of which being its \textbf{Programming Language}. 
A users' unfamiliarity with a specific language or its incompatibility with an existing pipeline might pose a burden for adoption.

Tied to the programming language are the \textbf{Compatible Packages} of a specific implementation. Compatibility with well-known packages greatly improves the usability of an implementation. Practitioners have also noted the compatibility with other packages as an important element when choosing a bias mitigation method \cite{Deng2022, Lee2021}.

\paragraph{Label Information}
\begin{itemize}
	\item \textbf{Programming Language.} States the programming language in which the method is provided. 
	\item \textbf{Compatible Package.} List of packages for which the compatibility with the method implementation has been shown. These can include data related packages such as pandas \cite{pandas2020}, folktables \cite{Ding2021}, or model related packages like scikit-learn \cite{scikit_learn}, Tensorflow \cite{Tensorflow} or PyTorch \cite{PyTorch}. 
	\item \textbf{Description.} The description includes the limitations of the implementation compared to the method's theoretical capabilities. It further includes practical information for integrating the method into a pipeline, such as the expected data format and hyperparameter information. 
\end{itemize}

\subsection{Use case}
The performance of a method is an important aspect in determining its usability. Performance is often shown through \textbf{Use Cases} which show the performance in a wide range of applications. As these use cases are strongly connected to their datasets, we report the datasets on which the method was tested. The description section allows providing initial performance parameters of the method. 

Recent work showed that the popular Adult, COMPAS, and German Credit datasets are not appropriate to benchmark bias mitigation methods \cite{Fabris_Messina_Silvello_Susto_2022, Ding2021}. The use case labels might help other researchers find appropriate sources. 

\paragraph{Label Information}
\begin{itemize}
	\item \textbf{Use Case.} The use case labels report the datasets on which the method has been tested, indicating its performance with regards to different tasks. An extensive overview of fairness datasets for a range of different tasks can be found in \citet{Fabris_Messina_Silvello_Susto_2022}.
	\item \textbf{Description.} The description allows reporting the performance of the method on its use cases, indicating its capabilities and allowing for easy comparison with other methods.
\end{itemize}

\section{Ongoing challenges}
Having detailed the structure and content of BiMi Sheets in the previous section, let us now discuss two ongoing challenges: the robustness of attempting to partially standardize labels describing bias limitation methods, and the sheets' adoption in practice. Note that these challenges are not unique to BiMi Sheets but hold for most fairness documentation initiatives.

\paragraph{Standardizing language} BiMi Sheets' structure aims to find a balance between harmonizing the language used to describe bias mitigation methods and retaining enough flexibility to account for various types of methods across different domains of machine learning. 

In particular, building on taxonomies and terms introduced in surveys, the sheets' proposed structure uses a set of labels to homogenize the lexicon used to characterize the properties of bias mitigation methods. The use of a common set of labels eases method comparison, and could help people less familiar with the broader field of AI fairness in understanding its intricacies.

However, we acknowledge that evolution in the field of AI fairness may require adjusting the labels used in BiMi Sheets. For instance, the introduction of new bias mitigation methods (e.g. the rise of methods providing some explainability in tandem with bias mitigation) may require the adjustment of method type labels; or the range of fairness type labels might require expansion as research attempts to account for a wider range of philosophical fairness notions.

\paragraph{Adoption of BiMi Sheets}
Besides providing a structured approach for documenting the properties of bias mitigation methods, the value of BiMi Sheets lies in standardizing the communication of these properties, and generally easing method comparisons. Fully realizing this potential would require their widespread adoption by both researchers and practitioners. 

As a first step towards that goal, we have created a platform to host BiMi Sheets, which can be found at bimisheet.com. It is populated with 24 BiMi Sheets, and allows the creation of novel sheets to be included in the platform after an acceptance process. The platform includes search capabilities and facilitates the comparison of hosted methods. This platform and the underlying sheets are fully open-source, allowing for other hosting initiatives.
In the Appendix you can find an offline version of the BiMi Sheets currently available on the platform.  

Besides adoption from individual researchers or practitioners, adoption from fairness frameworks such as AIF360 \cite{aif360_2018} and Fairlearn \cite{weerts2023fairlearn} would be beneficial for both parties. The goal of BiMi Sheets is complementary to that of fairness frameworks. This means that integrating with BiMi Sheets could improve the findability of methods provided by fairness frameworks, but also increase the coverage of BiMi Sheets, improving on its usefulness.

\section{Conclusion}\label{sec:discussion}

Motivated by previous research highlighting the challenges that the diversity of bias mitigation methods poses to both researchers and practitioners, this paper sought to tackle the problem of documentation debt in bias mitigation methods. BiMi Sheets propose to \emph{document} and \emph{structure} bias mitigation methods, by systematically reporting a method's description, pipeline constraints, fairness formalization, implementation details and use cases, empowering researchers and practitioners alike to better understand whether a method is applicable to their context.



Nevertheless, it should be emphasized that \emph{we did not `solve' the portability trap in fairness}. Documentation and structure alone cannot replace the holistic, interdisciplinary analysis required to understand how bias is best addressed in a decision process. Indeed, a technical solution may not necessarily be required, as also argued by \citet{Selbst2019} who discuss four other `traps'; and several works point to the inherent limitations of technical solutions to address biases in AI \cite{wachter2021fairness,john2022reality,buyl2024inherent}.

\bibliography{BiMiSheets}
\newpage
\appendix
\section{BiMi Sheet examples}
\label{appendix:examples}
On the following pages we provide 24 offline examples of BiMi Sheets. These methods either originate from well-known fairness toolkits or were proposed recently in top ML conferences. These choice criteria resulted in 16 BiMi Sheets from methods in AIF360 \cite{aif360_2018}, 4 from FairLearn \cite{weerts2023fairlearn}, 1 for error-parity \cite{cruz2024unprocessing}, 2 for fairret \cite{Buyl_Defrance_Bie_2024}, and 1 for OxonFair \cite{Delaney_Fu_Wachter_Mittelstadt_Russell_2024}.  

\clearpage



\section*{Supplementary material}

# Adversarial Debiasing: BiMi Sheet

## Metadata
Name: Adversarial Debiasing
Authors: AIF360
Version: 0.6.1    License: Apache 2.0
Proposed in *Mitigating Unwanted Biases with Adversarial Learning* [3]

## Method Description
( Adversarial Learning ) ( Regularization ) ( Binary Classification ) ( Tabular Datasets )

Adversarial debiasing is an in-processing technique that learns a classifier to maximize prediction accuracy and simultaneously reduce an adversary's ability to determine the protected attribute from the predictions. This approach leads to a fair classifier as the predictions cannot carry any group discrimination information that the adversary can exploit.

## Pipeline Architecture
( In-Processing ) ( Neural Networks )

The method is compatible with any gradient-based learning model, however the current implementation only allows for a neural network with one hidden layer.

## Fairness Type
( Binary Attribute ) ( Tunable Fairness Strength )
( Group Fairness ) ( Demographic Parity )

The adversarial model tries to predict the sensitive attributes from the prediction of the model. The method tries to prevent this from happening, effectively making the prediction independent of the sensitive attribute, which is equivalent to Demographic Parity.

## Implementation Constraints
( Python )

The model runs a Tensorflow neural network with one hidden layer. Input data has to be formatted as an AIF360 dataset.

1

### Tested Use Cases

<span style="background-color: #cccccc">Adult [1]</span>

Additional experiments were done in Zhang et al. [3] on the Google analogy data set [2].

# Calibrated Equalized Odds Post-processing: BiMi Sheet

## Metadata

Name: Calibrated Equalized Odds Post-processing
Authors: AIF360
Version: 0.6.1      License: Apache 2.0
Proposed in *On fairness and calibration* [4]

## Method Description

(Calibration) (Binary Classification) (Dataset Independent)

Equalized Odds Post-Processing is a post-processing technique that seeks to attain scores that optimize (a relaxation of) equalized odds while the classifier remains calibrated. This is achieved by changing the probabilities of certain samples to the base rate of the group rather than the classifier's prediction.

## Pipeline Architecture

(Post-Processing) (Calibrated Classifier)

Calibrated Equalized Odds Post-processing is compatible with any underlying learner that outputs a calibrated classifier with regards to the sensitive groups.

## Fairness Type

(Binary Attribute) (No Fairness Guarantee)
(Group Fairness) (Equal Opportunity) (Predictive Equality) (Calibration)

The method aims to satisfy either the chosen fairness measure, while maintaining calibration between satisfiers. This is achieved by sacrificing performance for one group. Experimental results show no guarantee on the performance of the method.

## Implementation Constraints

(Python) (scikit-learn)

While the concept of the method is compatible with any calibrated classifier, this implementation uses a set model type. The dataset must be passed in the AIF360 used format.

### Tested Use Cases

 Adult [1]   German [2]   Compas [3] 

The method does not seem to improve the fairness of a model when experimented on Adult.

# Deterministic Reranking - Conservative: BiMi Sheet

## Metadata

Name: Deterministic Reranking - Conservative
Authors: AIF360
Version: 0.6.1     License: Apache 2.0
Proposed in *Fairness-Aware Ranking in Search & Recommendation Systems with Application to LinkedIn Talent Search* [1]

## Method Description

`Re-ranking` `Ranking` `Dataset Independent`

If there are any groups for which the minimum representation constraint is violated, choose the element with the highest score among those groups. Otherwise, among groups that do not violate the maximum constraint, pick the group that minimizes $\frac{\lceil p_a * k \rceil}{p_a}$. From this group, choose the element with the highest score.

## Pipeline Architecture

`Post-Processing` `Score-based ranker`

Deterministic Reranking takes as input a list of output scores (for a each member of a list of candidates) of a ranker. Candidates are by convention assumed to be ordered by decreasing score by the initial ranker.

## Fairness Type

`Categorical Attributes` `Fairness Guaranteed`
`Group Fairness` `Adjusted Demographic Parity`

The Deterministic Reranker allows to set a desired proportion $p_a$ per demographic group $a$. Given k samples, the algorithm ensures that the number of selected candidates from each group $n_a$ satisfies $\lfloor p_a * k \rfloor \leq n_a \leq \lceil p_a * k \rceil$.

## Implementation Constraints

`Python` `scikit-learn`

The dataset must be passed in the AIF360 used format.

## Tested Use Cases

Synthetic Dataset

The original paper also tested the method on a proprietary dataset.

# Deterministic Reranking - Constrained: BiMi Sheet

## Metadata

Name: Deterministic Reranking - Constrained
Authors: AIF360
Version: 0.6.1     License: Apache 2.0
Proposed in *Fairness-Aware Ranking in Search & Recommendation Systems with Application to LinkedIn Talent Search* [1]

## Method Description

( Re-ranking ) ( Ranking ) ( Dataset Independent )

Starting with 0, increase the value of k until the minimum representation constraint is increased for at least one group. If there are more than one such groups, order them according to the descending score of their highest-scoring candidates not yet in the ranking. For each group in the list above: 1. Insert the next candidate from the group to the next empty index in the ranking Swap the candidate towards earlier indices until: Either the score of the candidate in the earlier index is larger, or, 2. Swapping will violate the minimum condition for the group of the candidate in the earlier index.

## Pipeline Architecture

( Post-Processing ) ( Score-based ranker )

Deterministic Reranking takes as input a list of output scores (for a each member of a list of candidates) of a ranker. Candidates are by convention assumed to be ordered by decreasing score by the initial ranker.

## Fairness Type

( Categorical Attributes ) ( Fairness Guaranteed )
( Group Fairness ) ( Adjusted Demographic Parity )

The Deterministic Reranker allows to set a desired proportion $p_a$ per demographic group $a$. Given k samples, the algorithm ensures that the number of selected candidates from each group $n_a$ satisfies $\lfloor p_a * k \rfloor \leq n_a \leq \lceil p_a * k \rceil$.

## Implementation Constraints

( Python ) ( scikit-learn )

The dataset must be passed in the AIF360 used format.

### Tested Use Cases
Synthetic Dataset

The original paper also tested the method on a proprietary dataset.

# Deterministic Reranking - Greedy: BiMi Sheet

## Metadata

Name: Deterministic Reranking - Greedy
Authors: AIF360
Version: 0.6.1   License: Apache 2.0
Proposed in *Fairness-Aware Ranking in Search & Recommendation Systems with Application to LinkedIn Talent Search* [1]

## Method Description

Re-ranking   Ranking   Dataset Independent

The attribute value with the highest next score among those that have not yet met their desired proportion is chosen. However, if there are any attribute values for which the desired proportion is about to be violated, the one with the highest next score among them is chosen.

## Pipeline Architecture

Post-Processing   Score-based ranker

Deterministic Reranking takes as input a list of output scores (for a each member of a list of candidates) of a ranker. Candidates are by convention assumed to be ordered by decreasing score by the initial ranker.

## Fairness Type

Categorical Attributes   Fairness Guaranteed
Group Fairness   Adjusted Demographic Parity

The Deterministic Reranker allows to set a desired proportion $p_a$ per demographic group $a$. Given k samples, the algorithm ensures that the number of selected candidates from each group $n_a$ satisfies $\lfloor p_a * k \rfloor \leq n_a \leq \lceil p_a * k \rceil$.

## Implementation Constraints

Python   scikit-learn

The dataset must be passed in the AIF360 used format.

## Tested Use Cases

Synthetic Dataset

The original paper also tested the method on a proprietary dataset.

# Deterministic Reranking - Relaxed: BiMi Sheet

## Metadata

Name: Deterministic Reranking - Relaxed
Authors: AIF360
Version: 0.6.1     License: Apache 2.0
Proposed in *Fairness-Aware Ranking in Search & Recommendation Systems with Application to LinkedIn Talent Search* [1]

## Method Description

(Re-ranking) (Ranking) (Dataset Independent)

If there are any groups for which the minimum representation constraint is violated, choose the element with the highest score among those groups. Otherwise, among groups that do not violate the maximum constraint, pick the group that minimizes $\lceil \frac{\lceil p_a * k \rceil}{p_a} \rceil$. From this group, choose the element with the highest score.

## Pipeline Architecture

(Post-Processing) (Score-based ranker)

Deterministic Reranking takes as input a list of output scores (for a each member of a list of candidates) of a ranker. Candidates are by convention assumed to be ordered by decreasing score by the initial ranker.

## Fairness Type

(Categorical Attributes) (Fairness Guaranteed)
(Group Fairness) (Adjusted Demographic Parity)

The Deterministic Reranker allows to set a desired proportion $p_a$ per demographic group $a$. Given k samples, the algorithm ensures that the number of selected candidates from each group $n_a$ satisfies $\lfloor p_a * k \rfloor \leq n_a \leq \lceil p_a * k \rceil$.

## Implementation Constraints

(Python) (scikit-learn)

The dataset must be passed in the AIF360 used format.

## Tested Use Cases

Synthetic Dataset

The original paper also tested the method on a proprietary dataset.

# Disparate Impact Remover: BiMi Sheet

## Metadata

Name: Disparate Impact Remover
Authors: AIF360
Version: 0.6.1    License: Apache 2.0
Proposed in *Certifying and Removing Disparate Impact* [3]

## Method Description

`Transformation` `Binary Classification` `Tabular Datasets`

Disparate Impact Remover changes the values of the labels while preserving ranking.

## Pipeline Architecture

`Pre-Processing` `Model Independent`

The original paper tested with Logistic Regression, Support Vector Machines, and Gaussian Naïve Bayes models.

## Fairness Type

`Binary Attribute` `Tunable Fairness Strength`
`Group Fairness` `Balanced Error Rate Parity`

The method aims to have the balanced error rate higher than a specific threshold. This follows

## Implementation Constraints

`Python` `scikit-learn`

The implementation of the method allows only binary attributes, the original paper states that the method can be applied for parallel attributes. Input data has to be formatted as an AIF360 dataset

## Tested Use Cases

`Adult [2]`

Experiments were conducted on the Ricci Data [1], German credit [4], and Adult income [2] in Feldman et al. [3].

1

# Grid Search Reduction: BiMi Sheet

## Metadata

Name: Grid Search Reduction
Authors: AIF360
Version: 0.6.1     License: Apache 2.0
Proposed in *A Reductions Approach to Fair Classification* [1]

## Method Description

( Constraint Optimization ) ( Binary Classification ) ( Dataset Independent )

Exponentiated gradient reduction is an in-processing technique that reduces fair classification to a sequence of cost-sensitive classification problems, returning a randomized classifier with the lowest empirical error subject to fair classification constraints.

## Pipeline Architecture

( In-Processing ) ( Model Independent )

Grid Search Reduction is compatible with any underlying learner that can produce 0-1 predictions.

## Fairness Type

( Categorical Attributes ) ( Fairness Guaranteed ) ( Group Fairness )
( Demographic Parity ) ( Equalized Odds ) ( Equal Opportunity )
( Predictive Equality )

A difference bound is passed to the system which details the allowed fairness violation. The difference between two groups will be at most twice the value of this difference bound hyperparameter.

## Implementation Constraints

( Python ) ( scikit-learn )

Input data has to be formatted as an AIF360 dataset. Users are required to provide a classifier or regression model satisfying scikit-learn conventions for the 'fit' and 'predict' methods (wth 0-1 outputs for 'predict').

## Tested Use Cases

( Adult [2] )

# GerryFair Classifier: BiMi Sheet

## Metadata

Name: GerryFair Classifier
Authors: AIF360
Version: 0.6.1 License: Apache 2.0
Proposed in *Preventing Fairness Gerrymandering: Auditing and Learning for Subgroup Fairness* [2] and *An Empirical Study of Rich Subgroup Fairness for Machine Learning* [3]

## Method Description

(Subgroup Analysis) (Calibration) (Binary Classification) (Tabular Datasets)

The algorithm is based on a formulation of subgroup fairness as a zero-sum game between a Learner (the primal player) and an Auditor (the dual player). The Learner is trying to minimize the sum of the prediction error and a fairness penalty term (given by the Lagrangian), and the Auditor is trying to penalize the fairness violation of the Learner by first identifying the subgroup with the greatest fairness violation and putting all the weight on the dual variable corresponding to this subgroup.

## Pipeline Architecture

(In-Processing) (Linear Regression) (Support Vector Machines)
(Decision Trees) (Kernel Regression)

## Fairness Type

(Categorical Attributes) (Fairness Guaranteed) (Group Fairness)
(Demographic Parity) (Equal Opportunity) (Predictive Equality)

The method stops when the difference in the subgroup's metric with the overall value of the metric weighted with their proportion is smaller than a chosen hyperparameter gamma.

## Implementation Constraints

(Python) (scikit-learn)

## Tested Use Cases

(Adult [1])

# Grid Search Reduction: BiMi Sheet

## Metadata
Name: Grid Search Reduction
Authors: AIF360
Version: 0.6.1     License: Apache 2.0
Proposed in *A Reductions Approach to Fair Classification, Fair Regression: Quantitative Definitions and Reduction-based Algorithms* [1, 2]

## Method Description
(Constraint Optimization) (Binary Classification) (Hard Labels) (Regression) (Dataset Independent)

Grid Search Reduction reduces the task to a sequence of cost-sensitive classification or regression problems, returning the deterministic classifier/regressor with the lowest empirical error subject to fairness constraints.

## Pipeline Architecture
(In-Processing) (Model Independent)

Grid Search Reduction is compatible with any underlying learner that can produce 0-1 predictions in the classification case, and any regression model in the regression case.

## Fairness Type
(Categorical Attributes) (No Fairness Guarantee) (Group Fairness) (Demographic Parity) (Equalized Odds) (Error Rate Parity) (True and False Positive Rate Parity) (Bounded Group Loss) (Moment based fairness definitions)

The approach applies to fairness definitions covered by *fairlearn.reductions.Moment* (including demographic parity, equalized odds, error rate parity, true and false positive rate parity for classification; and bounded group loss for regression).

## Implementation Constraints
(Python) (scikit-learn) (aif360)

Input data has to be formatted as an AIF360 dataset. Users are required to provide a classifier or regression model satisfying scikit-learn conventions

1

for the 'fit' and 'predict' methods. In the classification case, 'predict' should yield 0-1 outputs.

**Tested Use Cases**

Adult [3]    Law School [4]

# LFR: BiMi Sheet

## Metadata
Name: LFR
Authors: AIF360
Version: 0.6.1  License: Apache 2.0
Proposed in *Learning Fair Representations* [4]

## Method Description
(Adversarial Learning) (Binary Classification) (Tabular Datasets)

LFR maps each individual, represented as a data point in a given input space, to a probability distribution in a new representation space. The aim of this new representation is to lose any information that can identify whether the person belongs to the protected subgroup, while retaining as much other information as possible.

## Pipeline Architecture
(Pre-Processing) (Model Independent)

## Fairness Type
(Binary Attribute) (Tunable Fairness Strength)
(Group Fairness) (Individual Fairness) (Demographic Parity)

The method minimizes an objection which is a combination of demographic parity, information loss in the representation, and accuracy loss of the prediction label. The strength for each of these elements can be tuned.

## Implementation Constraints
(Python)

Input data has to be formatted as an AIF360 dataset.

## Tested Use Cases
(Adult [1])

Experiments were conducted on Adult [1], German credit [3], and Heritage Health Prize milestone 1 challenge [2] in Zemel et al. [4].

# MetaFair Classifier: BiMi Sheet

## Metadata

Name: MetaFair Classifier
Authors: AIF360
Version: 0.6.1 License: Apache 2.0
Proposed in *Classification with Fairness Constraints: A Meta-Algorithm with Provable Guarantees* [2]

## Method Description

`Constraint Optimization` `Binary Classification` `Tabular Datasets`

The MetaFair Classifier reduces classification with linear-fractional constraints to solving a small number of linear classification problems for carefully chosen parameters.

## Pipeline Architecture

`In-Processing` `Bayesian model`

The method has an accompanying model implementation used for predictions.

## Fairness Type

`Categorical Attributes` `Tunable Fairness Strength`
`Group Fairness` `Demographic Parity` `Predictive Parity`

The original model can satisfy any linear-fractional statistic, however only two are implemented here. The difference in fairness measure is noted as a fraction of the measures.

## Implementation Constraints

`Python` `scikit-learn`

Input data has to be formatted as an AIF360 dataset.

## Tested Use Cases

`Adult [1]`

Experiments on Adult [1], German credit [3] and COMPAS [4] were conducted in the original paper of Celis et al. [2].

# Optimized Preprocessing : BiMi Sheet

## Metadata
Name: Optimized Preprocessing
Authors: AIF360
Version: 0.6.1    License: Apache 2.0
Proposed in *Optimized Pre-Processing for Discrimination Prevention* [2]

## Method Description
 Transformation   Binary Classification   Tabular Datasets 

Optimized Preprocessing is a preprocessing technique that learns a probabilistic transformation that edits the features and labels in the data with group fairness, individual distortion, and data fidelity constraints and objectives.

## Pipeline Architecture
 Pre-Processing   Model Independent 

The original paper conducted experiments with Logistic Regression and Random Forest. [2]

## Fairness Type
 Categorical Attributes   Fairness Guaranteed 
 Group Fairness   Demographic Parity 

The fairness guarantee is provided on the dataset, not the model outcome. The method ensures that the fraction between positive rates is smaller than a hyperparameter epsilon, while keeping the distortion on the features under a given threshold.

## Implementation Constraints
 Python   scikit-learn 

The implementation does not use the initialized privileged and unprivileged groups in the initialization but the info on demographic attributes in the pre-processing of the dataset. The Optimizer dict contains the tunable parameters of the function.

## Tested Use Cases
 Adult [1]   German [3]   COMPAS [4]

# Prejudice Remover: BiMi Sheet

## Metadata

Name: Prejudice Remover  
Authors: AIF360  
Version: 0.6.1    License: Apache 2.0  
Proposed in *Fairness-aware classifier with prejudice remover regularizer* [2]

## Method Description

`Regularization` `Binary Classification` `Dataset Independent`

The Prejudice Remover adds a penalty to the learning objective that aims to reduce 'indirect prejudice' - the mutual information between the protected attribute and the model's predictions.

## Pipeline Architecture

`In-Processing` `Logistic Regression`

This bias mitigation method is currently only implemented for logistic regression.

## Fairness Type

`Binary Attribute` `Tunable Fairness Strength`  
`Group Fairness` `Demographic Parity`

The penalization terms aims to reduce 'indirect prejudice', minimizing mutual information between model predictions and sensitive attributes.

## Implementation Constraints

`Python`

Input data has to be formatted as an AIF360 dataset.

## Tested Use Cases

Experiments on Adult [1] were conducted in Kamishima et al. [2], it is not certain if this is the same implementation.

# Reject Option Classification: BiMi Sheet

## Metadata

Name: Reject Option Classification
Authors: AIF360
Version: 0.6.1     License: Apache 2.0
Proposed in *Decision theory for discrimination-aware classification* [3]

## Method Description

(Thresholding) (Binary Classification) (Hard Labels) (Dataset Independent)

Reject option classification is a postprocessing technique that gives favorable outcomes to unpriviliged groups and unfavorable outcomes to priviliged groups in a confidence band around the decision boundary with the highest uncertainty. This is done by flipping the labels for the samples of these groups close to the decision boundary.

## Pipeline Architecture

(Post-Processing) (Probabilistic Classifier)

Reject Option Classification is compatible with any underlying learner that can produce scores of predicted probabilities.

## Fairness Type

(Binary Attribute) (No Fairness Guarantee) (Group Fairness)
(Demographic Parity) (Equalized Odds) (Equal Opportunity)

The critical region size is determined based on optimizing the chosen fairness guarantee.

## Implementation Constraints

(Python) (scikit-learn)

Input data has to be formatted as an AIF360 dataset.

## Tested Use Cases

(Adult [1]) (German [2]) (Compas [4])

1

# Reweighing: BiMi Sheet

## Metadata
Name: Reweighing
Authors: AIF360
Version: 0.6.1  License: Apache 2.0
Proposed in *Data Preprocessing Techniques for Classification without Discrimination* [3]

## Method Description
(Reweighing) (Binary Classification) (Tabular Datasets)

The tuples in the training dataset are assigned weights. By carefully choosing the weights, the training dataset can be made discrimination-free w.r.t. S without having to change any of the labels. The weights on the tuples can be used directly in any method based on frequency counts.

## Pipeline Architecture
(Pre-Processing) (Model which incorporates weigths)

The method is effectively model independent, however for the method to be useful then the model must be able to incorporate the weights produced by the method.

## Fairness Type
(Binary Attribute) (No Fairness Guarantee)
(Group Fairness) (Demographic Parity)

The reweighing factor is the fraction of the expected frequency to the observed frequency wrt the combination of sensitive attribute and its class.

## Implementation Constraints
(Python) (scikit-learn)

Input data has to be formatted as an AIF360 dataset.

## Tested Use Cases
(Adult [1], German [4], COMPAS [2])

1

# Error-parity: BiMi Sheet

## Metadata

Name: Error-parity
Authors: André Cruz and Moritz Hardt
Version: 0.3.11    License: MIT License
Proposed in *Unprocessing Seven Years of Algorithmic Fairness.* [1]

## Method Description

( Thresholding ) ( Binary Classification ) ( Hard Labels ) ( Dataset Independent )

Error-parity sets groupspecific acceptance thresholds so as to **minimize risk while achieving an equality in error rates** across a desired set of groups. It is both simple and computationally efficient. Error-parity achieves exact error rate equality, unlike many preprocessing and inprocessing, which achieve some relaxation of the constraint. The method uses the output scores and returns hard prediction labels.

## Pipeline Architecture

( Post-Processing ) ( Probabilistic Classifier )

Error-parity is compatible with any underlying learner that can produce scores of predicted probabilities.

## Fairness Type

( Categorical Attributes ) ( Fairness Guaranteed ) ( Group Fairness )
( Demographic Parity ) ( Equalized Odds ) ( Equal Opportunity )
( Predictive Equality )

## Implementation Constraints

( Python 3.8-3.12 ) ( scikit-learn ) ( fairlearn )

The implementation requires a trained score predictor that takes in samples, X, in shape (num_samples, num_features), and outputs real-valued scores, R, in shape (num_samples,) as the model that feeds into error-parity.

## Tested Use Cases

( Synthetic ) ( Adult ) ( Folktables )

# Adversarial Mitigation: BiMi Sheet

## Metadata

Name: Adversarial Mitigation
Authors: Fairlearn
Version: 0.11.0       License: MIT
Proposed in *Mitigating Unwanted Biases with Adversarial Learning* [2]

## Method Description

(Adversarial Learning) (Classification) (Regression) (Tabular Datasets)

The predictor network is constructed to solve the underlying supervised learning task, without considering fairness, by minimizing the predictor loss. However, to improve fairness, we do not only minimize the predictor loss, but we also want to decrease the adversary's ability to predict the sensitive features from the predictor's predictions (when implementing demographic parity), or jointly from the predictor's predictions and true labels (when implementing equalized odds).

## Pipeline Architecture

(In-Processing) (Neural Networks)

The neural network models cannot have an activation function or discrete prediction function on the final layer.

## Fairness Type

(Categorical Attributes) (Numerical Attribute) (Tunable Fairness Strength)
(Group Fairness) (Demographic Parity) (Equalized Odds)

Fairness is achieved by minimizing the adversary's ability to predict the sensitive attribute. This is trade-off with the regular loss of the system, the strength of the fairness intervention can be tuned through a hyperparameter.

## Implementation Constraints

(Python) (scikit-learn) (PyTorch) (Tensorflow Keras)

The data needs to be provided as a 2D-arraylike of floats. The labels can be binary, categorical or float.

## Tested Use Cases

(Adult [1])

# Correlation Remover: BiMi Sheet

## Metadata
Name: Correlation Remover
Authors: Fairlearn
Version: 0.11.0      License: MIT

## Method Description
[ Transformation ] [ Task Independent ] [ Tabular Datasets ]

Sensitive features can be correlated with non-sensitive features in the dataset. By applying the CorrelationRemover, these correlations are projected away while details from the original data are retained as much as possible (as measured by the least-squares error).

## Pipeline Architecture
[ Pre-Processing ] [ Model Independent ]

We expect the CorrelationRemover to be most appropriate as a preprocessing step for (generalized) linear models.

## Fairness Type
[ Categorical Attributes ] [ No Fairness Guarantee ]
[ Dataset Unbiasing ] [ Removing Dataset Correlations ]

The degree to which the correlation is removed can be tuned with the $\alpha$ parameter.

## Implementation Constraints
[ Python ] [ scikit-learn ] [ PyTorch ] [ Tensorflow Keras ]

## Tested Use Cases
[ Diabetes 130-Hospitals [1, 2] ]

# Reductions: BiMi Sheet

## Metadata

Name: Reductions
Authors: Fairlearn
Version: 0.11.0      License: MIT
Proposed in *A reductions approach to fair classification.* [1] and
*Fair regression: quantitative definitions and reduction-based algorithms.* [2]

## Method Description

( Constraint Optimization )  ( Binary Classification )  ( Regression )
( Tabular Datasets )

A learning reduction takes as input complex examples, transforms them into simpler examples, invokes an appropriate learning algorithm on the simpler examples, then transforms predictions on these simpler examples to a prediction on the complex examples [3]. A fairness constraint is used to transform a binary classification or regression problem to a cost-sensitive problem.

## Pipeline Architecture

( In-Processing )  ( Probabilistic Classifier )

Reductions is compatible with any underlying learner that can produce scores of predicted probabilities.

## Fairness Type

( Categorical Attributes )  ( Fairness Guaranteed )  ( Group Fairness )
( Demographic Parity )  ( Equalized Odds )  ( Equal Opportunity )
( Predictive Equality )  ( Overall Accuracy Equality )  ( Bounded Group Loss )

A difference bound is passed to the system which details the allowed fairness violation. The difference between two groups will be at most twice the value of this difference bound hyperparameter.

### Implementation Constraints

`Python` `scikit-learn` `Tensorflow Keras`

The reduction algorithms in Fairlearn only require a wrapper access to any 'base' learning algorithm, meaning that the 'base' algorithm only needs to implement fit and predict methods, as any standard scikit-learn estimator, but it does not need to have any knowledge of the desired fairness constraints or sensitive features.

### Tested Use Cases

`Credit Card Clients [4]`

# Threshold Optimizer: BiMi Sheet

## Metadata
Name: Threshold Optimizer
Authors: Fairlearn
Version: 0.11.0     License: MIT
Proposed in *Equality of Opportunity in Supervised Learning* [2]

## Method Description
(Thresholding) (Binary Classification) (Hard Labels) (Dataset Independent)

For each sensitive feature value, ThresholdOptimizer creates separate thresholds and applies them to the predictions of the user-provided estimator. To decide on the thresholds it generates all possible thresholds and selects the best combination in terms of the objective and the fairness constraints.

## Pipeline Architecture
(Post-Processing) (Probabilistic Classifier)

ThresholdOptimizer expects an estimator that provides it with scores. While the output of ThresholdOptimizer is binary, the input is not limited to scores derived from binary classifiers. In fact, real valued input, e.g. from a regressor, provides it with many more options to create thresholds.

## Fairness Type
(Categorical Attributes) (Fairness Guaranteed)
(Group Fairness) (Demographic Parity) (Equalized Odds)
(Equal Opportunity) (Predictive Equality) (Predictive Parity)
(False Omission Rate Parity)

ThresholdOptimizer is built to satisfy the specified fairness criteria exactly and with no remaining disparity. In many cases this comes at the expense of performance, for example, with significantly lower accuracy. Regardless, it provides an interesting data point for comparison with other models.

## Implementation Constraints
(Python) (scikit-learn)

By default, ThresholdOptimizer trains the passed estimator using its fit() method. If prefit is set to True, ThresholdOptimizer does not call fit() on the estimator and assumes that it is already trained.

**Tested Use Cases**

Diabetes 130-Hospitals [1, 3]

# Fairret - Projection: BiMi Sheet

## Metadata
Name: Fairret - Projection
Authors: Maarten Buyl, MaryBeth Defrance, and Tijl De Bie
Version: 0.1.3    License: MIT
Proposed in *fairret: a Framework for Differentiable Fairness Regularization Terms* [1]

## Method Description
( Regularization ) ( Binary Classification ) ( Soft Labels ) ( Tabular Datasets )
( Image Datasets )

A FAIRRET quantifies a model's unfairness as a single value that is minimized like any other objective through automatic differentiation.

## Pipeline Architecture
( In-Processing ) ( Neural Networks )

The method is designed to be implemented in neural networks, however any method that uses gradients could be used.

## Fairness Type
( Parallel Attributes ) ( Numerical Attribute ) ( Tunable Fairness Strength )
( Group Fairness ) ( Demographic Parity ) ( Equal Opportunity )
( Predictive Equality ) ( Predictive Parity ) ( False Omission Rate Parity )
( Overall Accuracy Equality ) ( Treatment Equality ) ( F1-score Equality )

The method calculates the loss term by determining the distance of the model to the c-fixed fair set of models. Instead of using the linear-fractional models directly, the measures are set to the average value of the measure in a batch, allowing to transform them to linear constraints. These linear constraints constitute the fair set. The method can account for numerical attributes, however linear behavior is expected.

## Implementation Constraints
( Python ) ( PyTorch )

The package is implemented to work with PyTorch. No experiments were conducted to confirm the performance on image datasets.

**Tested Use Cases**

Bank [3]   CreditCard [4]   LawSchool (SEAPHE project)
ACS Datasets [2]

# Fairret - Violation: BiMi Sheet

## Metadata

Name: Fairret - Violation
Authors: Maarten Buyl, MaryBeth Defrance, and Tijl De Bie
Version: 0.1.3    License: MIT
Proposed in *fairret: a Framework for Differentiable Fairness Regularization Terms* [1]

## Method Description

Regularization   Binary Classification   Soft Labels   Tabular Datasets
Image Datasets

A FAIRRET quantifies a model's unfairness as a single value that is minimized like any other objective through automatic differentiation.

## Pipeline Architecture

In-Processing   Neural Networks

The method is designed to be implemented in neural networks, however any method that uses gradients could be used.

## Fairness Type

Parallel Attributes   Numerical Attribute   Tunable Fairness Strength
Group Fairness   Demographic Parity   Equal Opportunity
Predictive Equality   Predictive Parity   False Omission Rate Parity
Overall Accuracy Equality   Treatment Equality   F1-score Equality

A violation FAIRRET uses the proportion to which the fairness measure differs from the c-fixed measure as an added regularization term. This c-fixed measure is the average value for the measure within a batch, as this will be an integer it allows for the linear-fractional fairness measures to be transformed into linear equations. The method can account for numerical attributes, however linear behavior is expected.

## Implementation Constraints

Python   PyTorch

The package is implemented to work with PyTorch. No experiments were conducted to confirm the performance on image datasets.

**Tested Use Cases**

Bank [3]   CreditCard [4]   LawSchool (SEAPHE project)
ACS Datasets [2]

# OxonFair: BiMi Sheet

## Metadata
Name: OxonFair
Authors: E. Delaney, Z. Fu, S. Wachter, B. Mittelstadt, and C. Russell
Version: NA    License: Apache-2.0 license
Proposed in *OxonFair: A Flexible Toolkit for Algorithmic Fairness* [2]

## Method Description
(Thresholding) (Classification) (Hard Labels) (Inferring Sensitive Attributes)
(Dataset Independent)

OxonFair is a postprocessing approach for enforcing fairness, with support for a wide range of performance metrics and fairness criteria, and support for inferred attributes, i.e., it does not require access to protected attributes at test time. Under the hood, FairPredictor works by adjusting the decision boundary for each group individually. Where groups are not available, it makes use of inferred group membership to adjust decision boundaries.

## Pipeline Architecture
(Post-Processing) (Probabilistic Classifier)

Thresholding can be applied to most pretrained ML algorithms, and optimal thresholds can be selected using held-out validation data unused in training.

## Fairness Type
(Categorical Attributes) (No Attributes at Inference) (Fairness Guaranteed)
(Group Fairness) (Demographic Parity) (Equal Opportunity)
(Predictive Equality) (Predictive Parity) (False Omission Rate Parity)
(Overall Accuracy Equality) (F1-score Equality) (Min-Max)

OxonFair supports any fairness measure (including conditional fairness measures) that can be expressed per group as a weighted sum of True Positives, False Positives, True Negatives and False Negatives. OxenFair selects the solution that best optimizes the objective (performance) while satisfying the constraint (fairness bound).

## Implementation Constraints

`Python` `scikit-learn` `AutoGluon [4]` `PyTorch` `XGBoost [1]`

Oxonfair has its own dataloader object. A notebook is available detailing how to add a dataset not yet in OxonFair.

## Tested Use Cases

`COMPAS [9]` `CelebA [10]` `Multilingual Twitter corpus [7]` `Jigsaw [8]` `Myocardial Infarction [5]` `XGBoost [1]`

Oxonfair has shown to work with a Resnet-50 backbone [6] trained on ImageNet [3], decision trees, random forests, and XGBoost [1].



\end{document}